\documentclass[10pt,twocolumn,letterpaper]{article}

\usepackage{cvpr}
\usepackage{times}
\usepackage{tabularx}
\usepackage{booktabs}
\usepackage{epsfig}
\usepackage{graphicx}
\usepackage{amsmath}
\usepackage{amssymb}
\usepackage{lib}
\usepackage{cite}
\usepackage[english]{babel}
\usepackage{blindtext}
\usepackage{lib2}
\usepackage[keeplastbox,nospread]{flushend}
\usepackage[font=footnotesize,labelfont=bf]{caption}
\AtBeginDocument{
  \addtolength{\abovedisplayskip}{-1ex}
  \addtolength{\abovedisplayshortskip}{-1ex}
  \addtolength{\belowdisplayskip}{-1ex}
  \addtolength{\belowdisplayshortskip}{-1ex}
}

\newcommand\blfootnote[1]{%
  \begingroup
  \renewcommand\thefootnote{}\footnote{#1}%
  \addtocounter{footnote}{-1}%
  \endgroup
}

\usepackage[pagebackref=true,breaklinks=true,letterpaper=true,colorlinks,bookmarks=false,citecolor=blue]{hyperref}
\cvprfinalcopy 

\def\cvprPaperID{3.14159265359} 

\begin{document}

\renewcommand{\paragraph}[1]{\par \noindent {\bf #1}}

\title{Unifying Map and Landmark Based Representations for Visual Navigation}

\author{Saurabh Gupta $\;$ David F. Fouhey $\;$ Sergey Levine $\;$ Jitendra Malik\\
UC Berkeley \\
{\tt\small \{sgupta, dfouhey, svlevine, malik\}@eecs.berkeley.edu}}

\maketitle
\thispagestyle{empty}

\blfootnote{Project website with videos:
\url{https://s-gupta.github.io/cmpl/}}

\begin{abstract}
This works presents a formulation for visual navigation that unifies map based
spatial reasoning and path planning, with landmark based robust plan execution
in noisy environments. Our proposed formulation is learned from data and is
thus able to leverage statistical regularities of the world. This allows it to
efficiently navigate in novel environments given only a sparse set of
registered images as input for building representations for space. Our
formulation is based on three key ideas: a learned path planner that outputs
path plans to reach the goal, a feature synthesis engine that predicts features
for locations along the planned path, and a learned goal-driven closed loop
controller that can follow plans given these synthesized features. We test
our approach for goal-driven navigation in simulated real world environments
and report performance gains over competitive baseline approaches.  
\end{abstract}

\section{Introduction}
As humans we are able to effortlessly navigate through environments. We can
effectively plan paths, find shortcuts and adapt to changes in the
environment while only having a relatively coarse sense of the exact
geometry of the environment or the exact egomotion over long distances.

Human representations for space \cite{wiener2009taxonomy} start out from
\textit{destination knowledge}, which is a set of places of interest. This guides
acquisition of \textit{route knowledge} that is knowledge of specific routes to
go from one destination to another. Already at this point, humans exhibit an
impressive ability to follow routes reliably in the presence of 
changes in the environment (demolition and construction of new buildings,
appearance variations due to change in seasons, and the time of day)
as well as the uncertainty in their own estimate of their egomotion and location in
space. This is facilitated by the use of distinctive landmarks (such as clock towers) that are invariant
to changes in the environment and provide visual fixes
to collapse the uncertainty in location. 
As more and more of such route knowledge is experienced over time, it is
assimilated into \textit{survey knowledge}, which is knowledge of the spatial
layout of different places and routes with respect to one another. It is at
this point that humans are also able to plan paths to novel places and find
shortcuts for going between different known points. This is generally
accomplished by reasoning over a coarse spatial layout of places through a
cognitive map \cite{tolman1948cognitive}.

\begin{figure}
\insertWL{1.0}{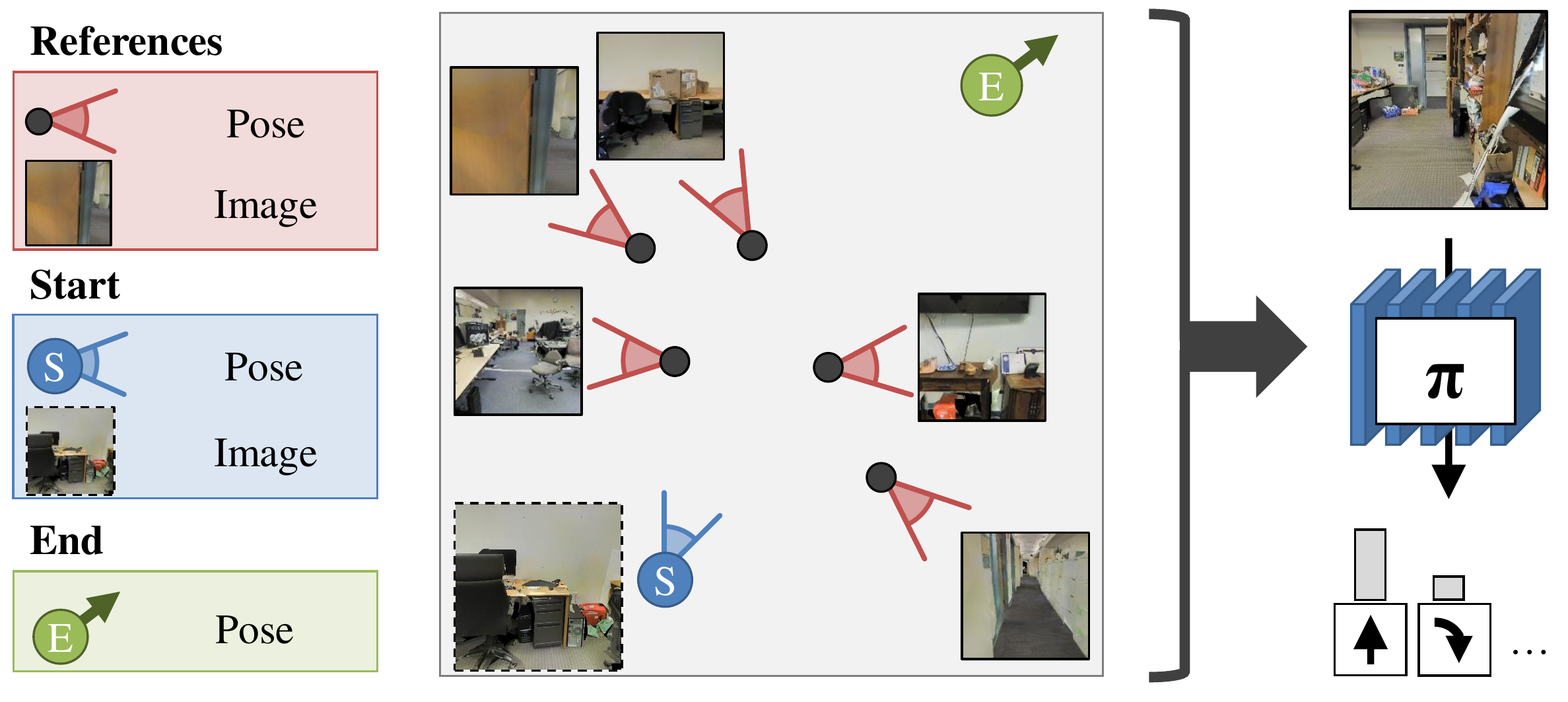}
\caption{\textbf{Problem Setup}: Given a set of reference images and poses,
the starting image and pose, and a goal pose, we want a policy $\pi$ 
that is able to convey the robot from its current pose to the target pose 
using first person RGB image observations under noisy actuation.}
\figlabel{setup}
\vspace{-0.1in}
\end{figure}

This remarkably human navigation ability is driven by both {\it map
based} and {\it landmark based} representations. Map based representations
allow for planning paths and finding shortcuts, while landmark based visual
memory of the environment permits executing the planned path in the presence of
noisy and uncertain motion. At the same time, humans use their prior experience with other similar
environments to construct reasonable path plans with only partial
knowledge of a new environment.

Of course, there is vast literature in robotics that studies navigation.
Classical approaches employ purely geometric reasoning. This provides accurate
geometry, but does not explicitly reason about the semantic distinctiveness of
landmarks. At the same time, there is a strict requirement of being able to
{\it always} localize the agent in this map. Not only is this localization
solely driven by geometry, precise localization may not even be always
possible (such as in long texture-less hallways). More crucially, classical
approaches rely entirely on observed evidence and do not use priors and
experience from similar other environments to make meaningful inferences
about unobserved parts of the environment.

Contemporary learning based approaches \cite{zhu2016target,
mirowski2016learning} have focused on incorporating such statistical priors,
but have largely ignored the question of representation of spaces by making it
{\it completely} implicit in the form of neural network activations. Works that
do build spatial representations (such as \cite{gupta2017cognitive}), operate
under ideal world assumptions where the agent is always aware of its exact
location in the environment. 

In this work, we build map and landmark based representations for spaces: map
based representations are used to plan paths and find shortcuts, while past
visual memories are used to derive visual fixes (landmarks) to reliably follow
planned paths under noisy actuation. 
We start by generalizing the cognitive mapping and planning work from Gupta
\etal \cite{gupta2017cognitive}
to work with sparse distributed observations of an environment.  Next, instead of
explicitly localizing with respect to the map at each time step (to read out
the next action), we compile a given path plan into a `path signature',
comprising the model's expectation about what the agent will see along the
path in addition to the planned actions. 
This path signature, along with the current observation, is used with a learned
policy for closed-loop control to output actions that follow the planned path
robustly. If the robot doesn't see the features it expected, it can take
corrective actions without explicit relocalization or replanning.  While this
approach is quite feasible even with standard geometric mapping and planning
methods, our learning-based framework makes it particularly effective: we can
express path plans semantically (\eg, go through the door, turn at the end of
the corridor), as well as, learn to produce good path signatures for the end task of
robust path following.
Finally, all parts of our representation are learned from data in an end-to-end
manner for the sub-tasks of path planning and path following. This allows us to
use prior experience with similar other environments to make meaningful
extrapolations, allowing our policies to operate well even with only partial
observation of novel environments.

Our experiments in simulated environments (based on reconstructions of real
world offices) with simple discrete actions, demonstrate the role of past
visual memories for robust plan execution in presence of noisy actuation, and
the utility of our specialized policy architectures over standard memory-based
neural network architectures.

\section{Related Work}
There are multiple different branches of research in classical and learning
based robotics that tackle the problem of robot navigation. We describe some
major efforts here.
Robotic navigation research can be broadly categorized along two dimensions:
planning-based versus reactive, and learning-based versus geometric. Many of
the most popular classical method are planning-based and geometric, and rely on
separating mapping and estimation from decision making. Indeed, these two
problems are often studied in isolation in robotics
\cite{siciliano2016springer}, as we will discuss below. Most learning-based
methods for navigation have been reactive in nature, bypassing the explicit
construction of maps or planning. In contrast, our method is \emph{both}
learning-based and planning-based: we combine end-to-end training with explicit
construction of spatial representations and planning.

\paragraph{Classical Methods for Navigation.} 
There is rich literature in various aspects of navigation covering
construction of maps from sensory measurements like images and range scans,
localization in such maps through probabilistic state estimation, and path
planning with noisy actuation. We summarize a few works here, and a
comprehensive review appears in \cite{thrun2005probabilistic}.

Filtering and estimation methods deal with the problem of estimating the
robot's pose given a history of its sensor readings, typically in a known map
that is purely geometric.  Methods like particle filtering maintain a
probability distribution over the robot's location and update this distribution
based on a generative model of the sensors \cite{doucet2000rao}. Extending this
to the setting of unknown environments, simultaneous localization and mapping
(SLAM) algorithms \cite{davison1998mobile, slam-survey:2015,engel2014lsd,
klein2007parallel, izadiUIST11} aim to both construct a map and localize the
robot. These methods are solely focused on constructing maps that are purely
geometric and entirely ignore semantics. Odometry and localization methods
based on computer vision sometimes also aim to localize the robot without
explicitly constructing a map.  The most common class of such methods is visual
odometry \cite{nister2004visual, zhou2017unsupervised}, where the goal is to
estimate camera motion based on sequences of images in order to determine
displacement from a starting point. Standard robotic navigation methods
typically decouple mapping or localization from decision making. Once
localization has been performed, the action might be determined by a path
planning procedure. Many path planning methods \cite{canny1988complexity,
kavraki1996probabilistic, lavalle2000rapidly} assume direct access to a
noiseless map, and often a noiseless robot location. A number of methods also
incorporate uncertainty, by feathering the obstacles in the environment by the
maximum amount of uncertainty to obtain provably safe plans
\cite{axelrod2017provably}.

\begin{figure*}
\insertWL{1.0}{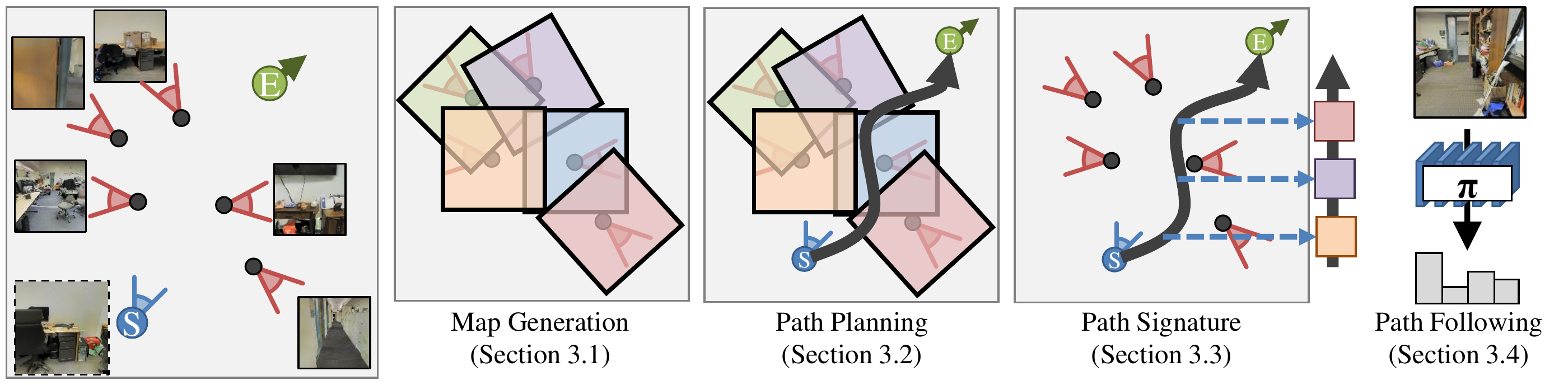}
\caption{\textbf{Overview}: Given a small set of registered \rgb images as input,
our approach for visual navigation conveys the robot to the target location using
two central components: a joint mapping and planning module that learns to
plan paths and find shortcuts given the sparse set of registered views; and a
joint path signature and execution module that executes a given path.}
\figlabel{overview}
\vspace{-0.1in}
\end{figure*}

Given such a constructed map and plan, robot motion happens using a control
loop that 1) executes the first action of the plan, 2) relocalizes the robot in
the map using new observations, 3) replans to compute optimal action from the
new state to the goal location. This control loop crucially depends on being
able to accurately relocalize the robot at every time step. Such accurate
relocalization is problematic when the environment is featureless (a long
texture less hallway) or if it changes (chairs are moved around); or if it
becomes necessary to explore new routes. Moreover, all steps in this control
loop are entirely based on geometry; and state estimation and action execution
are entirely decoupled. Being purely geometric makes this control loop
extremely brittle. Furthermore, decoupled estimation and execution makes it
impossible to adapt the perception and localization systems to the final task
performance. Our work addresses these issues by framing localization as an
implicit learning problem thereby allowing use of geometric as well as semantic
cues. At the same time, our control loop is entirely learned. This allows for
specialization of the implicit localization module for action execution. 

Researchers have also looked at view based maps \cite{konolige2010view} that
store views at different locations instead of geometric maps. Such methods
directly tie action execution to sensors without learning. This requires
maintain enormous collections of images and leads to policies that entirely
limited to previously traversed paths.

\paragraph{Learning based Navigation.}
Inspired by the short comings of pipelined approaches, there is an increasing
focus on end-to-end learned task-driven architectures for robotic applications
\cite{levine2016end}. Most such approaches \cite{pomerleau1989alvinn,
sadeghi2016cadrl, kahn2017self, giusti2016machine, daftry2016learning} study
task-agnostic collision avoidance, and don't build representations for spaces
or plan paths.
Even task-driven approaches such as \cite{zhu2016target, mirowski2016learning,
brahmbhatt2017deepnav} either rely only on statistical priors for spaces via
learned reactive policies or only build implicit representations in the form of
vanilla LSTM memory. Such implicit representations fail to quickly adapt to new
environments.
In contrast to the above methods, techniques based on learned map construction
\cite{gupta2017cognitive, parisotto2017neural, khan2017memory, zhang2017neural} 
and value iteration networks \cite{tamar2016value}
have sought to directly incorporate elements of mapping and planning into
end-to-end trained navigational policies. These prior methods are closest to
our approach. However, they only study the simplified problem without any noise
in actuation. This entirely side-steps the problem of localization in the map.
We adopt their general paradigm of mapping and planning but extend it to work
with only a small set of sparse views. More
importantly, we study the problem in the more challenging setting where there
is actuation noise. This limits the utility of maps and plans, and we propose
techniques to tackle this by unifying landmark based visual fixes to compensate
for noisy in actuation.

Another related line of work is learning from demonstration
\cite{schaal1997learning} and imitation learning where a policy is trained to
follow a given demonstration or observation sequence. Such approaches rely on
an external expert to demonstrate the desired behavior at test time and
additionally  also require actual or close to actual observations that will be
encountered while executing the trajectory. In contrast, our work neither
requires expert demonstrations nor actual exact images that will be seen when
we execute the trajectory.

\section{Approach}
\begin{figure*}
\insertWL{1.0}{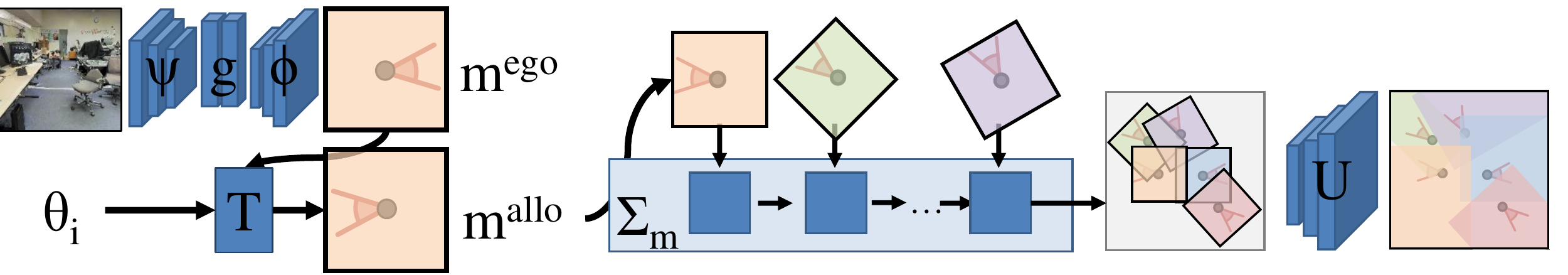}
\caption{\textbf{Mapping Architecture}: Given an input image and pose $\theta_i$, 
our mapping architecture passes it through a neural network composed of 
$\psi$, $g$, and $\phi$, producing an egocentric map $m^{ego}$ from the camera view. 
This map is transformed according to $\theta_i$ using a spatial transformer to produce
an allocentric map $m^{allo}$ in the world frame. The allocentric maps are accumulated
by $\Sigma_m$ and upconvolved by $U$ to produce the final map.}
\figlabel{mapplan}
\vspace{-0.1in}
\end{figure*}

In this work, we are interested in building representations for spaces starting
from visual observations. Let us consider an embodied agent that has some small
amount of experience with a novel environment. This experience is in the form a
small number (say $N$) of registered \rgb images ($I_i$) along with associated
poses ($\theta_i$). We denote this set of registered images by $\mathcal{I}$.
Given this experience with the environment, we want to build representations
for this environment that enable this agent to quickly accomplish navigation
goals under noisy actuation (\ie the actions executed by the agent don't
always have the exact intended effect). In particular, given current
observations from on-board sensors (\rgb image from first person camera) and
the compiled representation for space, we want the agent to get to a specified
point in space.

Our method is based on two main components: a) a spatial
representation that allows path planning and shortcut inference and b) a
goal-driven closed-loop controller that executes the plan under noisy
actuation. For the former, we adopt and generalize the learning based mapping
and planning paradigm from Gupta \etal \cite{gupta2017cognitive} to operate
with a small number of sparse views ($\mathcal{I}$) of the environment. 
For the latter, our key insight is that a plan should not just consist of a
sequence of actions that will reach the goal, but a `path signature' that
additionally also describes the features that the agent expects to see on the
way to the goal. This path signature can then be used for closed-loop control:
if the robot doesn't see the features it expected, the agent can use relatively
simple, local, error correction strategies to take corrective actions without
explicit relocalization and replanning. 

Our approach is summarized in \figref{overview} and consists of four parts: 
a) a \textit{learned mapper} that produces spatial representations of the world
given sparse views of the environment;
b) a \textit{learned path planner} that processes the produced map along with
the target position to predict optimal paths and guides the learning of the
mapper;
c) a \textit{learned feature synthesizer} that hallucinates features at
locations along the predicted path using the available sparse views from the
environment; and
d) a \textit{learned control policy} that executes the plan based on the
hallucinated features and the current first person observation under noisy
actuation.

Our approach is reminiscent of classic robotics pipelines, but with the crucial
distinction that, instead of using geometry-based, hand-defined, isolated
modules, we propose a differentiable, end-to-end trainable, data-driven
formulation.
This addresses some of the fundamental challenges with standard, geometric
localization and planning methods. Classical approaches are driven entirely by
explicit geometry and have no way to incorporate statistical priors. They also
require design of intermediate representations such as: What should the output
of the mapper be? What constitutes a landmark? What features will be useful for
matching the current observation to a previously observed landmark or how to
decide that a landmark has been reached. Such intermediate representations are
hard to design and even harder to optimize given surrogate metrics that do not
always correlate with the final performance.  Finally, independent modules are
usually unaware of the error statistics of the other modules and hence cannot
be robust to them.

In contrast, in our proposed formulation, learning allows use of statistical
priors based on knowledge of similar other environments; enabling our method to
work with only a handful of images.  Moreover, joint optimization of modules
alleviates the need to hand define intermediate representations that bottleneck
the final performance. Furthermore, modules can be trained together, allowing
them to be robust the error-modes of one another. We now describe the design of
each module.

\subsection{Map Generation}
\seclabel{map}
The task of our map generator is to transform the registered images to a spatial
map that can be used for path planning. Going from images to a common spatial
representation facilitates path planning and shortcut finding. Unlike a
classical mapper, our proposed mapper is not constrained to output free or
occupied space but only a representation that allows for effective path
planning. The only structure that we impose is that it is spatial
representation \ie locations in the map correspond to locations in the world.
The mapper is free to output arbitrary abstract feature vectors as long as they
allow a planner to output meaningful plans.

Given the registered image samples $\mathcal{I}$,
the mapping function $M$ generates the overhead map $m$ as follows:
\begin{eqnarray}
m^{ego}_i &=& \phi\left(g\left(\psi\left(I_i\right)\right)\right); \\
m^{allo}_i &=& T\left(m^{ego}_{i}, \theta_{i}\right); \\
m &=& U\left(\Sigma_m \left(m^{allo}_1, m^{allo}_2, \ldots, m^{allo}_N\right)\right).
\end{eqnarray}
$m^{ego}_i$ is generated using an encoder/decoder network that learns to
transform from the first person view to an egocentric top view of the
environment at location $\theta_i$.  The encoder $\psi$ is composed of
convolutional layers that map from raw image pixels to abstract representation
for the scene. This representation is transformed through fully connected
layers $g$ into a top-view that is upsampled by upconvolutional layers $\phi$
to produce feature vectors in an egocentric spatial top view of the world
($m^{ego}_i$) for each image $I_i$.  These egocentric top-view predictions
are transformed into a common allocentric view (\ie in the world coordinate
frame) through a geometric transform determined by the image's pose
$\theta_i$ to generate $m^{allo}_i$. This transform is implemented
differentiably using spatial transformers \cite{jaderberg2015spatial}.
Different allocentric predictions are accumulated by a function $\Sigma_m$
(realized through weighted averaging) and finally upsampled by function $U$
using deconvolutions. The map generator transforms the input image observations
into an allocentric representation of the world. This allocentric
representation of the world can be plugged into downstream modules and
optimized jointly.

\subsection{Path Planning} 
\seclabel{plan}
Given the abstract spatial map, the task of the planner is to output a path
(sequence of actions) that can convey the robot to a desired and known goal
location. We are working with abstract spatial representations 
(\ie not explicitly free-space), a
classical planner can no longer be used. Instead, we learn a planner from data. 
Given the map $M(\mathcal{I})$ in the top-view, we
employ a value iteration based planning method to plan a path from the starting
location to the goal location. We express value iteration as a
convolutional network with max pooling across channels \cite{tamar2016value}
and directly train the planner and the mapper to output the optimal
action at different locations in the map \cite{gupta2017cognitive}. We read off
the entire sequence of actions ($a^p_j$) along the path $p$ that convey the
robot to the goal location by following optimal actions from the starting
location to the goal location. This action sequence can be converted into a
sequence of robot poses $\rho^p_j$ in the environment.

Formulating mapping and path planning as learning problems allows us to fuse
observations of the current environment with statistical knowledge about
layout of similar other environments. This lets us meaningfully extrapolate and make
predictions about regions of space where no direct observations may have been
made. We can thus plan paths in new environments given only a handful of
images at sparse locations, and generate more efficient paths than are
possible from only the sparse observations.

\subsection{Path Signatures}
\seclabel{synth}
The sequence of planned actions ($a^p_j$ and $\rho^p_j$) is by itself not
enough to execute the trajectory in the environment because of noisy actuation:
actions executed by the agent will not lead to the same consistent outcome because
of environmental noise, slip in motors and uneven terrain. This makes
open loop replay of plans problematic because errors compound over time and the
location drifts: the agent will execute the optimal plan for the wrong location 
making its location error worse and worse.
Such drift in estimate of one's location also occurs in humans and animals as
they move around and is compensated for by use of visual fixes by re-orienting
oneself with respect to a previously known distinctive location.  We thus
compensate for drift by augmenting the sequence of actions and poses $\rho^p_j$
with visual anchors $\hat{f_j^p}$ at that location. These feature anchors
provide a reference to the agent allowing it to estimate and correct for its
drift. 

Unfortunately, we may not have explicitly been at $\rho^p_j$ in the past and
may not have the {\it actual} features $f^p_j$ for that location. To address this we
{\it synthesize} features $\hat{f^p_j}$ for location $\rho^p_j$. We do this
using observed
features from neighboring points $q$ in space that the agent did visit in the
past and thus has some features $f^q$ for it.

\begin{figure}
\insertWL{1.0}{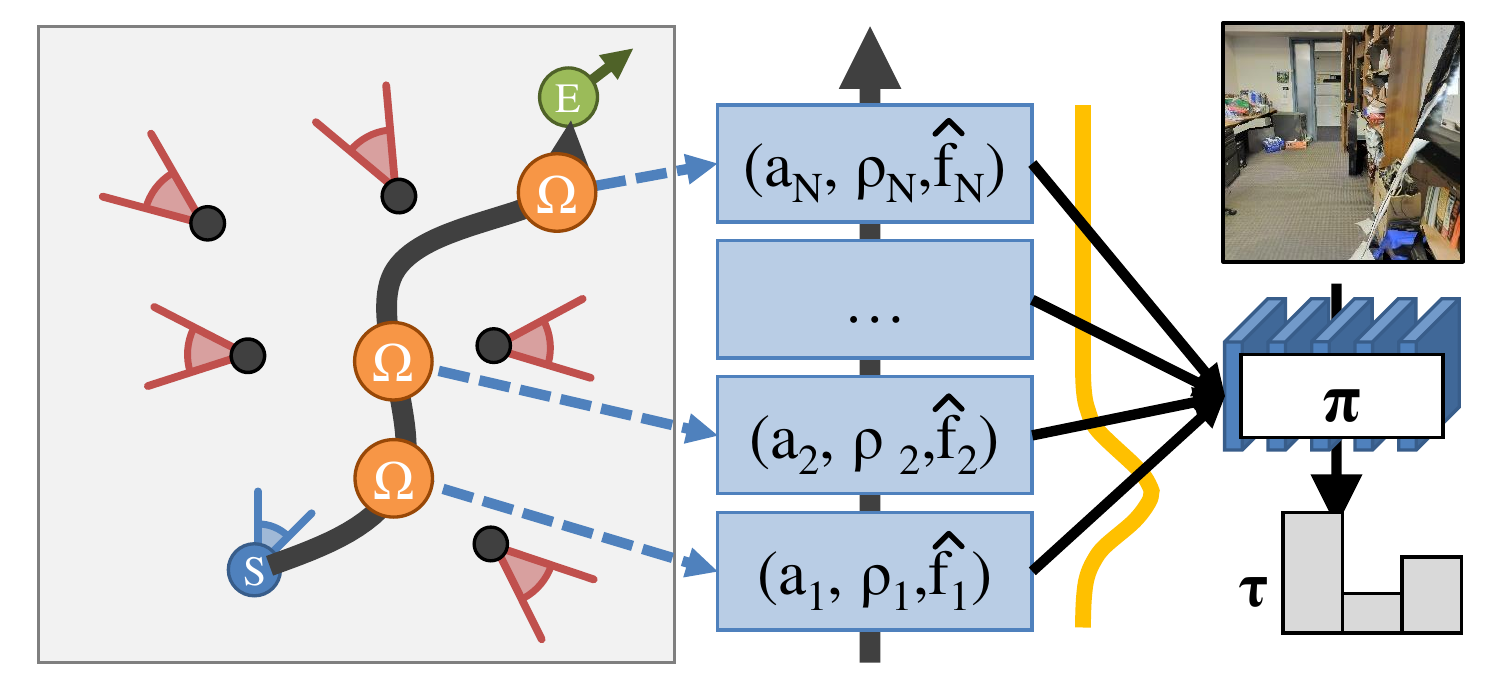}
\caption{\textbf{Feature Synthesis and Policy Execution}: 
Given a path, we produce a signature consisting of a sequence
of tuples $(a_i, \rho_i, \hat{f}_i)$., These tuples are softly attended
to by a path executing policy $\pi$ which in sequence determines
what action to take.}
\vspace{-0.1in}
\figlabel{execute}
\end{figure}

Thus for each location along the plan we have the action, the location and the
synthesized features \ie the tuple $(a^p_j, \rho^p_j, \hat{f^p_j})$.
We call the sequence of these tuples a signature $\Xi(p)$ of the path $p$ that
we want the robot to follow:
\begin{eqnarray}
\Xi(p) &=& \{(a^p_j, \rho^p_j, \hat{f^p_j}) : j \in [1 \ldots J]\}
\end{eqnarray}
We next describe how to synthesize features $\hat{f^p_j}$ for a location
$\rho^p_j$ using observation $\mathcal{I}$ from neighboring locations:
\begin{eqnarray}
\omega_{i,j} &=& \Omega\left(\left(\psi(I_i), \delta(\theta_i, \rho^p_j)\right)\right) \\
\hat{f^p_j} &=& \Sigma_f \left(\omega_{1,j}, \omega_{2,j}, \ldots, \omega_{N,j}\right) 
\end{eqnarray}
Here, the function $\delta$ computes the relative pose of image $I_i$ with
respect to the desired synthesis location $\rho^p_j$. $\psi$ computes
representation for image $I_i$ through a CNN followed by two fully connected
layers. $\Omega$ fuses the relative pose with the representation for the image
to obtain the contribution $\omega_{i,j}$ of image $I_i$ towards representation at
location $\rho^p_j$. These contributions $\omega_{i,j}$ from different images are
accumulated through a weighted addition by function $\Sigma_f$, to obtain the
synthesized feature $\hat{f^p_j}$ at pose $\rho^p_j$. This, as
well as the plan execution module $\pi$, is depicted in \figref{execute}.

\subsection{Plan Execution}
\seclabel{exec}
Given a path signature that has the local information needed to
follow the trajectory in the environment, we need a policy that can execute
this path under noisy actuation.

This learned policy takes the environment and goal specific path signature
$\Xi(p)$ as input. This factorization of the environment and goal specific
information into a path signature that is separate from the policy lets us learn
a \textit{single} policy that can do different things in different
environments with different path signatures without requiring any re-training
or adaptation. The policy can then also be
thought of as a robust closed-loop goal-oriented controller.

We train this policy $\pi$ to follow path $p$ under noisy actuation in the
environment given current first person views $o_t$ from on-board robot cameras.
We represent policy $\pi$ with a recurrent neural network that iterates over
the path signature $\Xi(p)$ as the agent moves through the environment. This
iteration is implemented using sequential soft attention that traverses
over the trajectory signature. At each step, the path signature is read into $\xi$ with
differentiable soft attention centered at $\eta$:
\begin{eqnarray}
\vspace{-0.05in}
\xi &=& \sum_j \varphi\left(\Xi(p)_j\right) e^{-|\eta_t-j|}. 
\vspace{-0.05in}
\end{eqnarray}
The recurrent function $\pi$ with state $\zeta_t$ is implemented as:
\begin{eqnarray}
\vspace{-0.05in}
\zeta_{t+1}, \eta, \tau &=& \pi(\zeta_t, \xi, \psi(o_t)), \\
\eta_{t+1} &=& \eta_{t} + \sigma(\eta).
\vspace{-0.05in}
\end{eqnarray}
As input, it takes the internal state, $\zeta_t$, attended path signature
$\xi$ and featurized image observation $\psi(o_t)$. In return, it gives
a new state $\zeta_{t+1}$, pointer increment $\eta$, and action $\tau$
that the agent should execute. This pointer increment is added, after a sigmoid
to yield the new pointer $\eta_{t+1}$. We set $\eta_1 = 1$ and $\zeta_1 = \mathbf{0}$.
Our experiments show that this use of sequential reasoning here is critical for performance.

\begin{figure*}
\insertH{0.22}{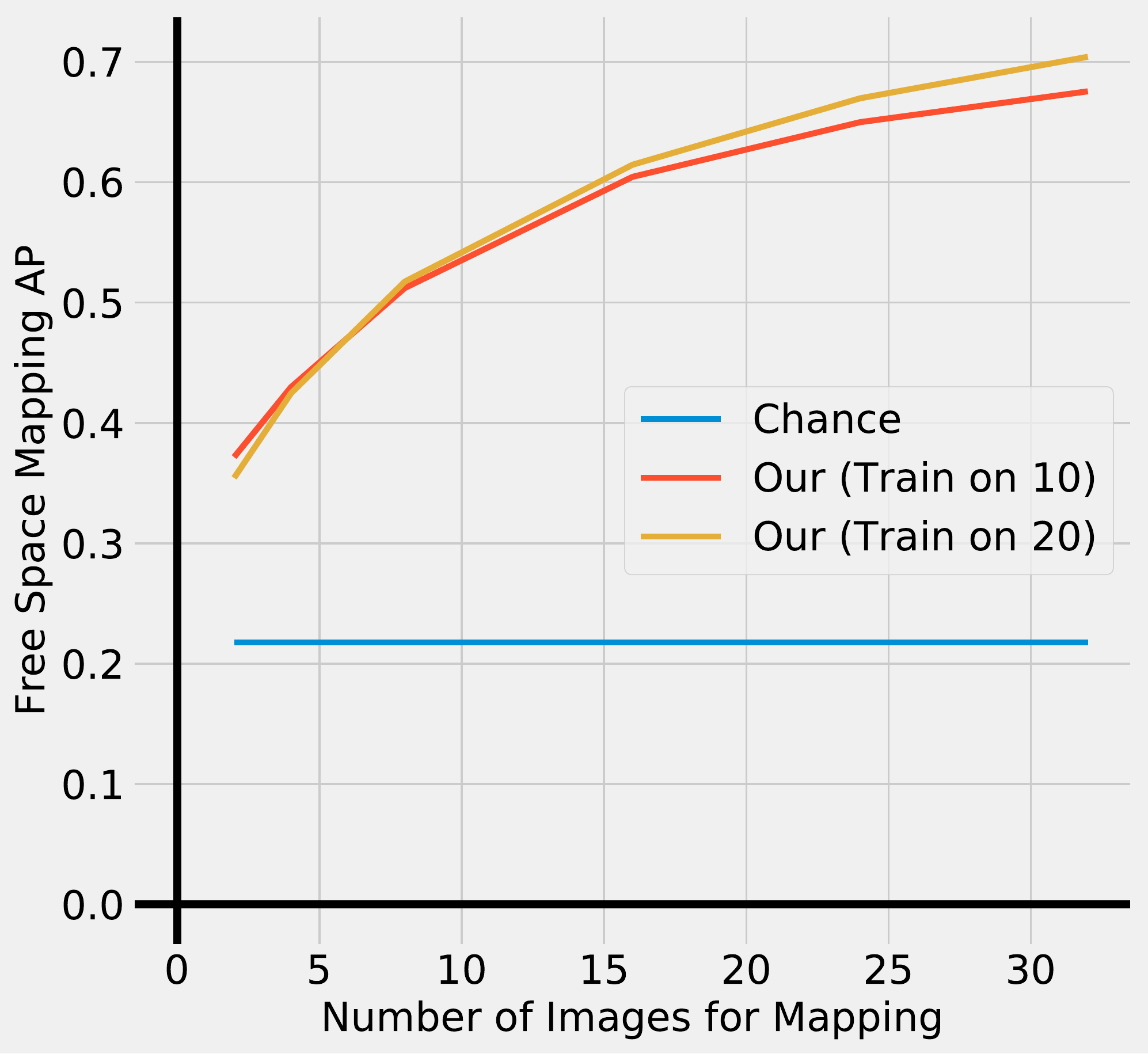} \hfill
\insertH{0.22}{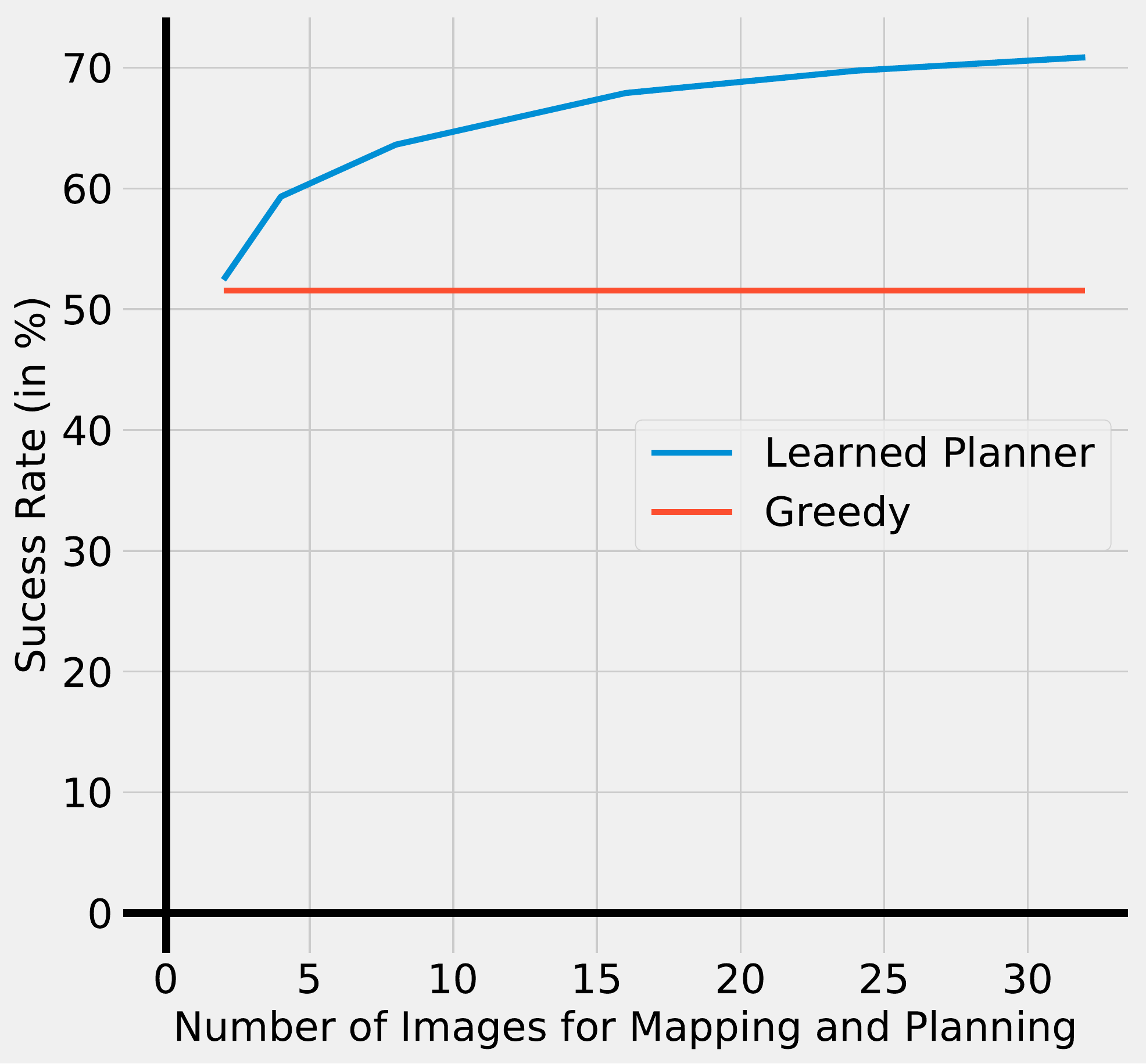} \hfill
\insertH{0.22}{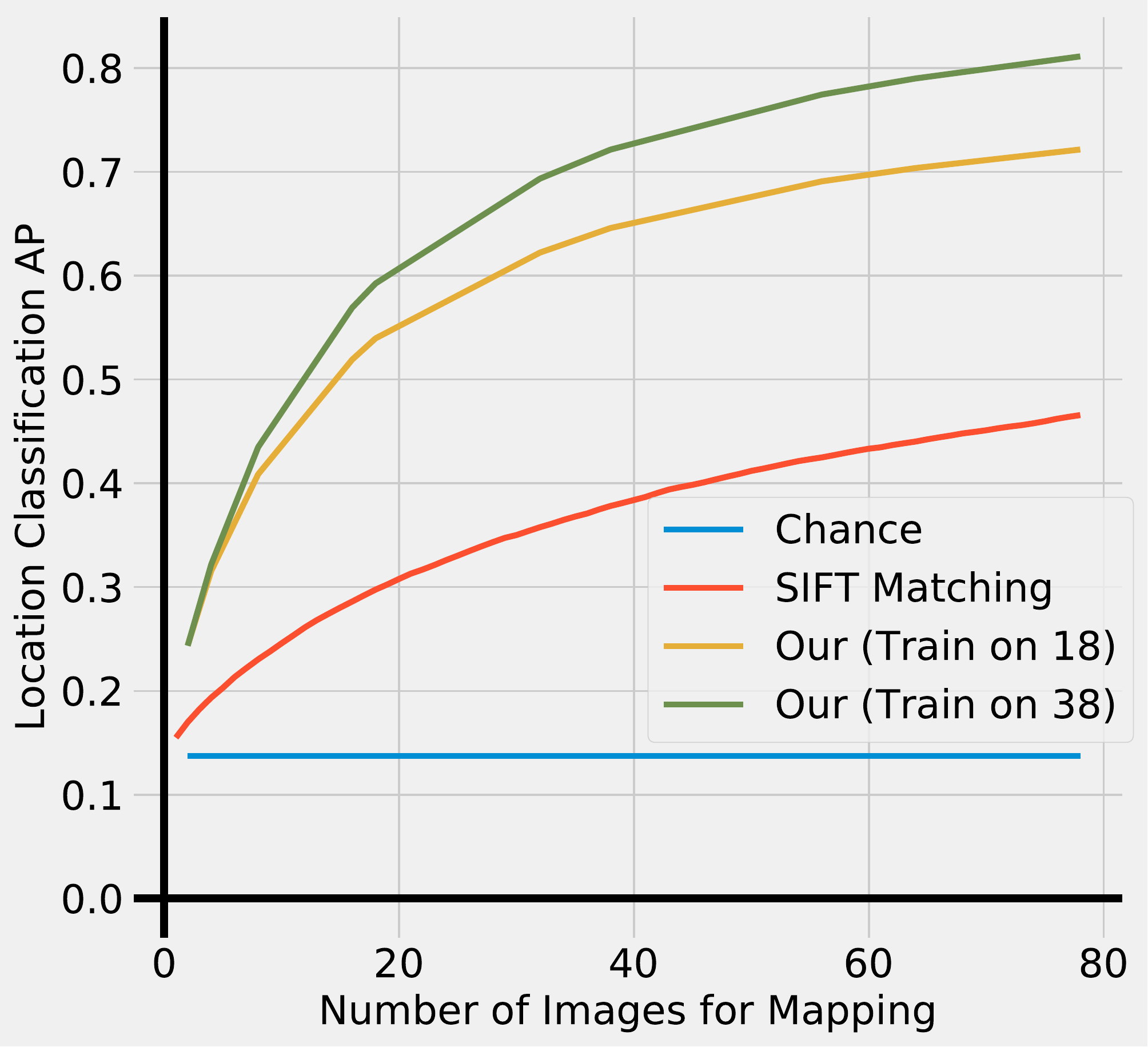} \hfill
\insertH{0.22}{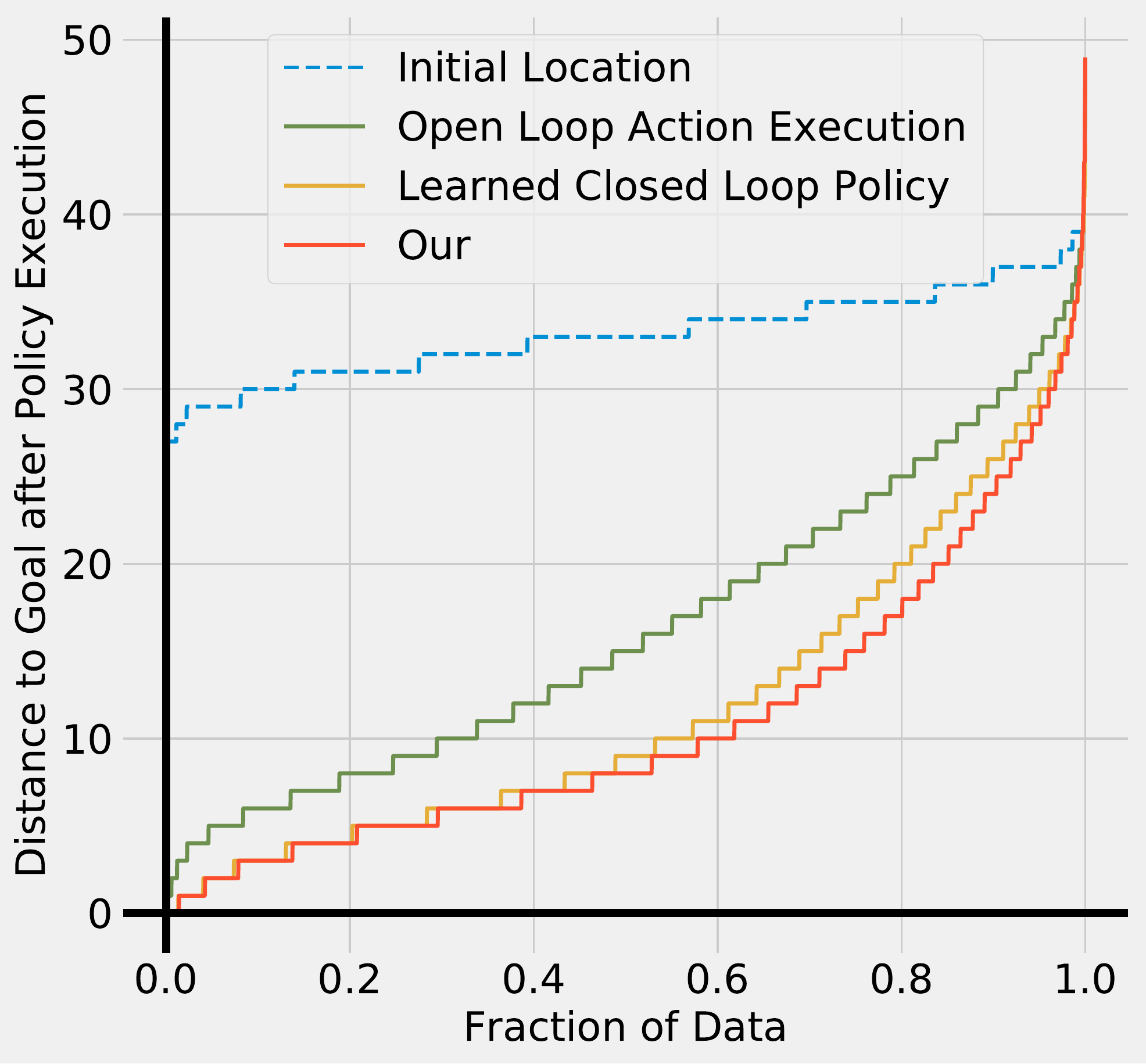}
\caption{\textbf{Results: }Far left: Average Precision for free space prediction 
as a function of the number of mapping images when the mapper module is trained in
isolation for the task of free space prediction. Center left: Success rate for getting 
to the goal using the joint mapper and planner architecture to output open loop
plans varying number of registered images as input. Center right: Average precision
for localization of novel locations in space given varying number of reference images
as input. Far right: Distribution of distance to goal after for following a plan to
get to a goal that is 32 to 36 time steps away with reference images coming randomly 
from around the trajectory. Best viewed in color, see relevant 
text for details.}
\figlabel{loc_ap}
\vspace{-0.1in}
\end{figure*}

\section{Experiments}
We describe experiments to evaluate our proposed representation for space for
the task of visual navigation.  We present a factored evaluation: we evaluate
the individual modules, and the two coupled modules (mapping and planning
modules together, and path signature and execution modules together). Such a
factored evaluation helps us evaluate the design choices for each module
without conflating the error modes of different modules. 

The goal of the experiment section is to answer the following questions: given
only a handful of registered images in a novel environment and under noisy
actuation: (1) how well mapping and planning work; (2) to what extent can
features for novel locations be predicted; (3) how well can a given trajectory
plan be executed; (4) whether visual memory is helpful; and (5) how different
architectures for visual memory compare against each other?

\paragraph{Experimental Setup.} Our goal is to enable robot mobility in indoor
environments. It is challenging to conduct thorough and controlled
experimentation with physical robots. Thus, we use environments that are
simulated, but are derived from reconstructions of real buildings (Matterport scans
\cite{matterport}).  These environments retain many of the challenges of the
real world, such as clutter and realistic appearance, but allow for controlled
experiment and systematic analysis. 

We use the Stanford Large-Scale 3D Indoor Spaces Dataset \cite{armeni20163d}.
This dataset contains 3D scans of multiple different buildings on the Stanford
campus, in the form of textured meshes that can be rendered from arbitrary
viewpoints, and has been used for studying navigation tasks
\cite{gupta2017cognitive}. We follow the standard splits for our experiments,
using areas 1, 5, 6 for training, area 3 for validation, and area 4 for
testing. This ensures that testing is done in a  {\it different} building from
ones used for training and validation.  We adapt the publicly available
simulation environment from Gupta \etal \cite{gupta2017cognitive} for our
experiments.

\begin{figure}
\setlength{\fboxsep}{0pt}
\setlength{\fboxrule}{0.2pt}
\fbox{\insertWL{0.235}{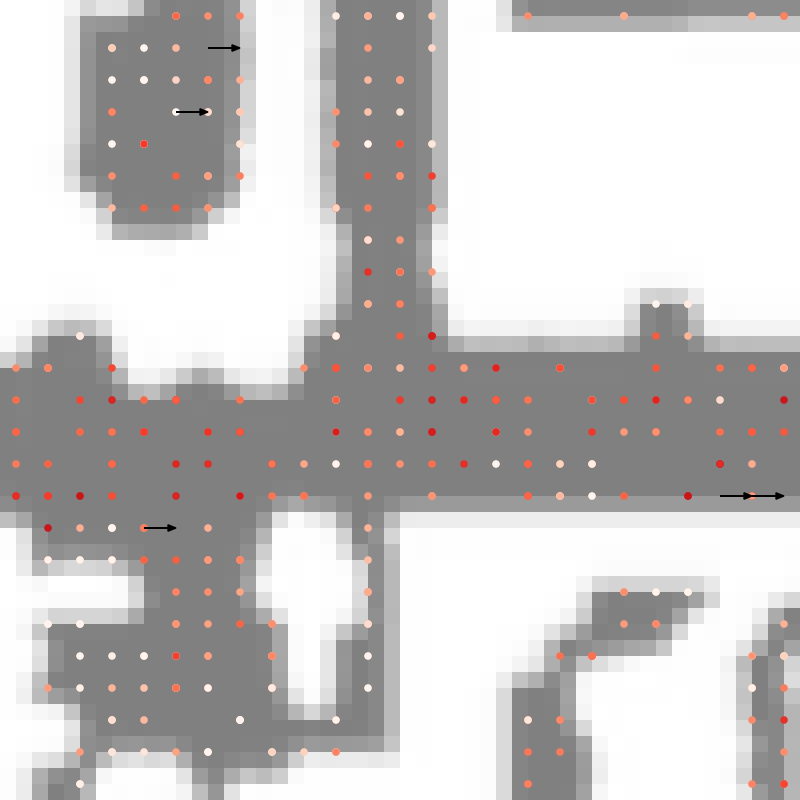}} \hfill
\fbox{\insertWL{0.235}{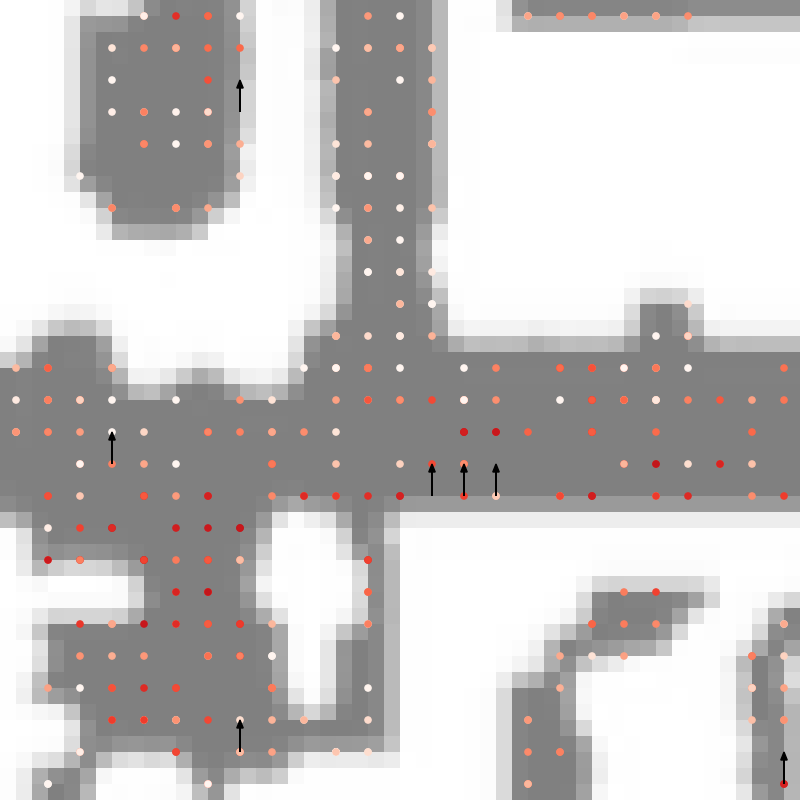}} \hfill
\fbox{\insertWL{0.235}{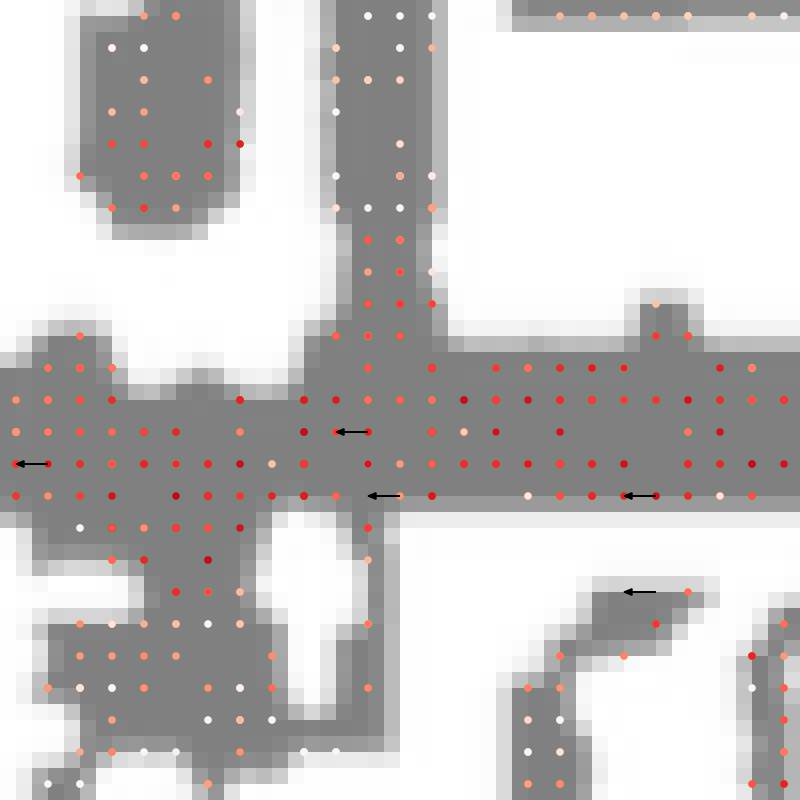}} \hfill
\fbox{\insertWL{0.235}{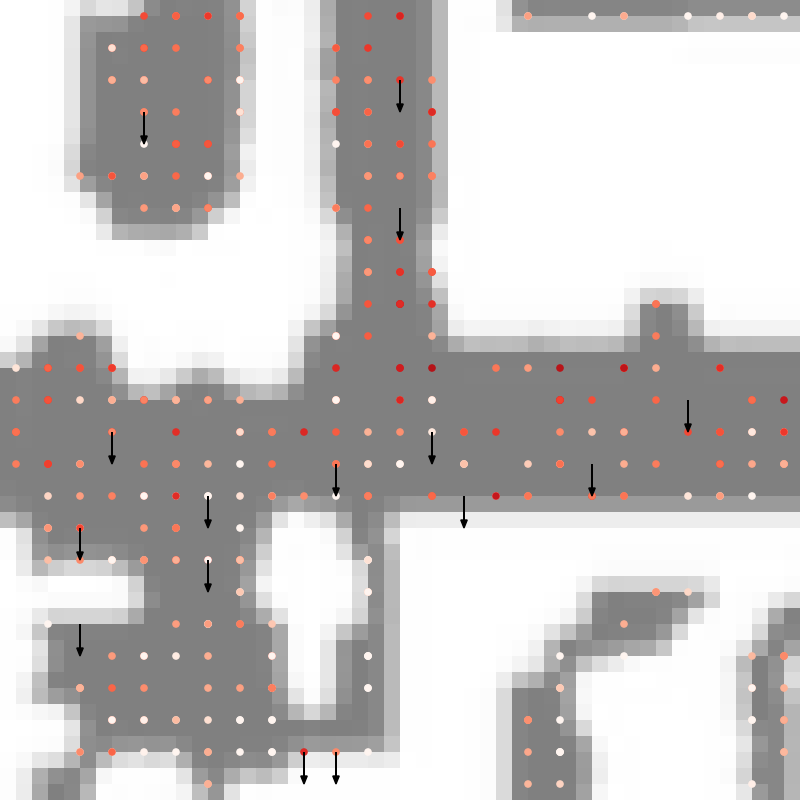}} \hfill
\caption{\textbf{Visualization for quality of feature synthesis}: We plot the 
quality of localization at different locations (denoted by red dots) 
in space given reference images (marked in black). The four plots show 
the four different orientations (facing right, facing up, facing left 
and facing down) and the color of the dot indicates the quality of localization 
(darker red indicates higher score). We see performance is good when
reference images in the same facing direction are available in close neighborhood.} 
\figlabel{synth_vis}
\vspace{-0.1in}
\end{figure}

\subsection{Mapping and Planning}
\paragraph{Mapping.} We first study how well our approach can
generate maps given a small number of images from a novel environment.  We
first study this in isolation by training the mapper described
in \secref{map} to predict free space in an environment. Note
that this experiment aims only to establish that the mapper architecture
is sufficiently expressive; in all other experiments, it is trained end-to-end
by backpropagating gradients from upstream modules.

We sample a small number of images with their pose over the area in
consideration ($32m\times32m$ in our experiments) and train the mapper to
predict free space over the entire area.  We benchmark the performance of this
mapper in the test environment and measure the average precision for the task
of free space prediction. Our mapper can take a varying number of images as
input; we plot performance as a function of the number of test-time images in
\figref{loc_ap}~(far left).  Even though the mapper was trained with a fixed
number of images ($10$ or $20$) as input, it works well when tested with
different numbers.

\paragraph{Mapping and Planning.} We next study how well paths can be planned
with sparse observations. We use the joint mapper and planner architecture
described in \secrefs{map}{plan} and train it for the task of going to a
particular point in space up to 20 steps away within areas of size
$16m\times16m$. As previously described, the mapper and planner generates a
spatial map and uses value iteration to output optimal actions to convey the
robot to the goal from different locations in the map.

Our goal here is to evaluate {\it only} the ability of this system to generate good 
plans from sparse observations. Therefore, we evaluate how well the plans 
can be open-loop executed under the assumption of no noise or feedback.
We randomly sample start and goal positions and measure the plan execution
rate if we simply apply the sequences of actions starting from the start state.
\figref{loc_ap} (center left) show the success rate as we vary the number of
reference images. Again we see performance improves as more and more images are
available from the environment.

It is important to note that these representations and path plans are obtained
from a small number of \rgb images with a limited field of view ($60^\circ$).
Classical methods rely on being able to compute correspondences between images
to reconstruct the world. In our setup, there is a very limited overlap between
the different images and such methods simply do not work.  Similar experiments
were reported by Gupta \etal in \cite{gupta2017cognitive}.  However, Gupta
\etal were studying the problem of online mapping and planning and thus, their
analysis was limited to streams of consecutive images along a path. Dense
observations lead them to develop an egocentric representations and permitted
re-planning as new observations came in. In contrast, our experiment here
studies the case where only sparse image observations are available. We thus
maintain an allocentric representation but show that good path plans can be
generated even with sparse observations of the environment. Additionally, in
this experiment we are studying the open loop performance of such plans. This
is in contrast to the experiments reported in Gupta \etal that only reported
the performance of closed loop plans.

\begin{figure*}
\setlength{\fboxsep}{0pt}
\setlength{\fboxrule}{0.2pt}
\centering
\insertWL{1.00}{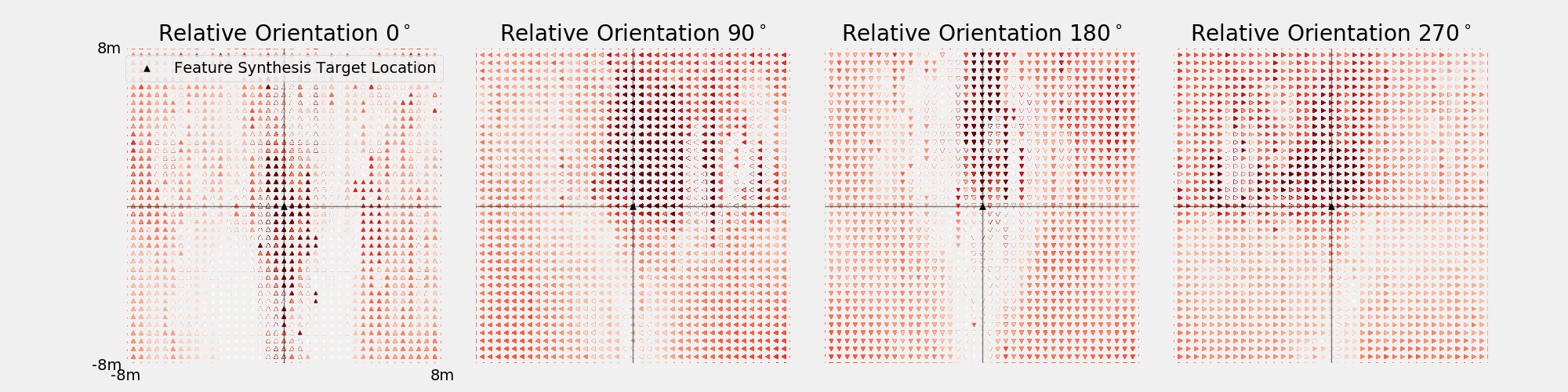}
\caption{Visualization for which reference locations contribute to the features at 
the origin in facing up orientation. We use the weights with which features from 
different reference images are combined as a proxy measure for the importance of 
that reference image for synthesizing features at the target location. We plot this measure
of importance over space (and orientation in different figures). Deeper reds indicate stronger 
contribution. Once again location along the line of sight of the target 
orientation contribute more. Interestingly images looking at the target location from front
also contribute strongly. Image facing perpendicularly also contribute weakly when there
is overlap in visual field of view.}
\figlabel{synth_vis_1}
\vspace{-0.1in}
\end{figure*}

\subsection{Path Signatures and Execution}
We now study feature synthesis for unseen locations from nearby
locations, and their use for robustly executing planned trajectories.

\paragraph{Feature Synthesis.} We first analyze feature synthesis for novel
points in isolation. This serves as a test bed to evaluate the expressiveness of
the feature synthesis architecture.

For this evaluation, we train the synthesizer to do localization at novel
locations: given the set of images $\mathcal{I}$ in an area ($32m\times32m$), we
predict features at different target locations using the architecture described
in \secref{synth}.  We train the synthesizer discriminatively for the task of
pairing synthesized features $\hat{f_\theta}$ at location $\theta \in
\Theta$ to actual features $f_\phi$ at location $\phi \in \Phi$. We use a
multi-layer perceptron classifier on $[\hat{f_\theta}; f_\phi]$ to predict if
they are from nearby locations (as determined by the distance between $\theta$
and $\phi$), for a number of different pairs of $\theta$ and $\phi$.

We report the average precision for this classification task in
\figref{loc_ap}~(center right). Note that this network can take a varying number
of registered reference images as input, and we plot the performance as a
function of the number of reference images. We compare to a SIFT matching
based baseline (checking consistency of SIFT keypoint matches with the
essential matrix derived from the relative pose between the two images). Our
learned features outperform the SIFT baseline.
\figref{synth_vis} show visualizations for the quality of feature prediction at
different locations for a representative prediction, and \figref{synth_vis_1}
shows the contribution of different locations towards predicting features at a
canonical position (at origin looking upwards).

\paragraph{Path Signatures and Plan Execution.}
We now evaluate the design of our architecture for executing planned
trajectories. In addition to the set of registered images ($\mathcal{I}$), we
are given a plan (in the form of sequence of actions and locations in space),
that we want the agent to successfully execute. To not conflate with errors in
planning, we conduct this experiment with optimal plans for reaching a target
point in space (sampled to be between 16 and 18 steps away). 

We make some additional modeling assumptions: we use a relatively simple
action representation (four discrete actions: stay in place, rotate left by
$90^\circ$, rotate right by $90^\circ$ and move forward by $40cm$). We
simulate noise in actuation by making the move forward action fails with $20\%$
probability. The agent stays in place when the action fails. 

\paragraph{Training Details.} We train our policy from \secref{exec}, using
\rldagger \cite{ross2011reduction}, an imitation learning algorithm. We
supervise the agent using the optimal action for reaching the goal from the
agent's actual current state (known in the simulator).  We found it useful to
include an auxiliary loss between the features $f^p_j$ of the actual
observation from location $\rho^p_j$ and the predicted features $\hat{f}^p_j$
at that location, and slowly switching from $f^p_j$ to $\hat{f}^p_j$ for
computing the path signatures as the training proceeded. We also found that
using a shallow 5-layer \cnn (with strided convolutions for down-sampling
instead of max pooling) to represent images led to better performance than
using an ImageNet pre-trained \resnet network. We believe strided convolutions
and slower down-sampling in our custom network eases learning of pose sensitive
representations for reasoning about view-point and geometric image similarity.
Policies were trained using \adam \cite{kingma2014adam} for $60K$ iterations,
and learning rate was dropped every $20K$ iterations. 

\paragraph{Metrics.} Performance is measured by executing all policies for $20$
steps and measuring the distance to the goal at end of the execution.
\tableref{policy-eval} reports the mean distance to goal,
75\textsuperscript{th} percentile distance to goal, and the success rate
(finishing within 3 steps of the goal) over $8K$ such episodes.

\paragraph{Baselines.} We compare to the following baselines.

\par \noindent \textit{Open Loop Execution:} This policy is an open loop
execution of the planned sequence of actions. It ignores the input image and
merely applies the planned action sequence, as is, one action at a time.
\par \noindent \textit{Learned Action Execution:} Our next comparison point is a
policy that is trained to execute the planned trajectory in the environment.
This policy has exactly the same architecture as our proposed architecture
(\secref{exec}) except that the path signature is derived only from the
sequence of planned actions (and not the associated synthesized views). At run
time, this policy receives the current image observation and the planned action
sequence and it outputs the action that the agent should execute in the
environment. It can use the feedback from the environment (in form of the new
image observation) to adapt to failed actions due to noisy actuation. This is
a strong baseline. It can learn sensible patterns such as: it is more
likely to turn when facing an obstacles, and it is more likely to turn after
exiting a room rather than before exiting the room.

\begin{figure*}
\setlength{\fboxsep}{0pt}
\setlength{\fboxrule}{0.2pt}
\insertH{0.25}{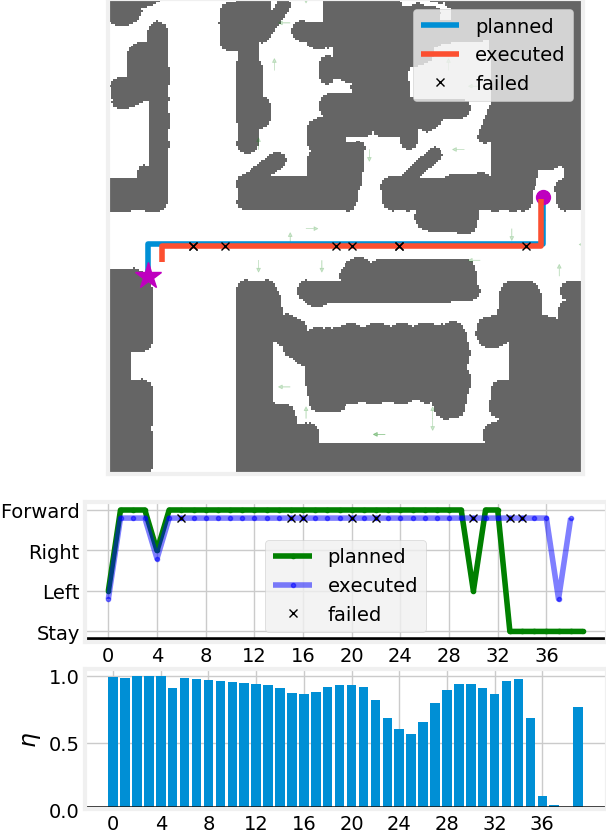} \hfill
\insertH{0.25}{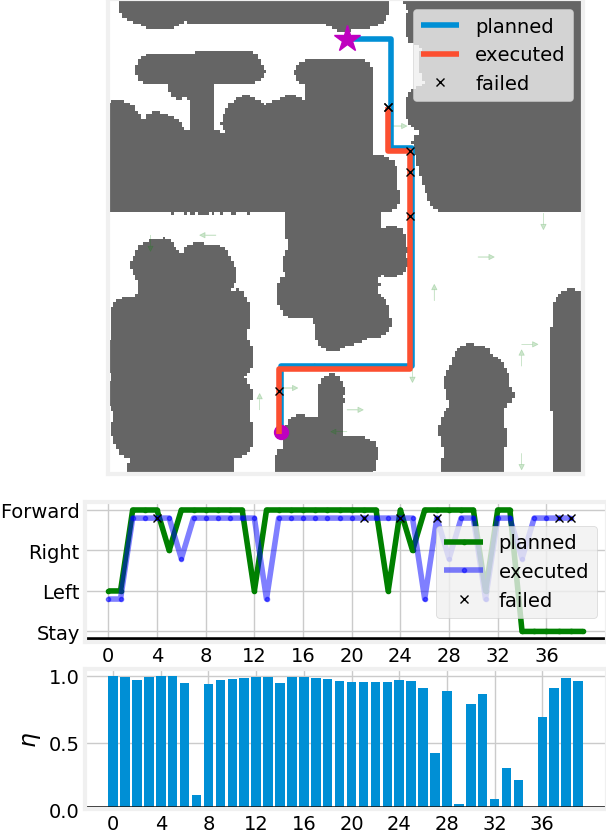} \hfill
\insertH{0.25}{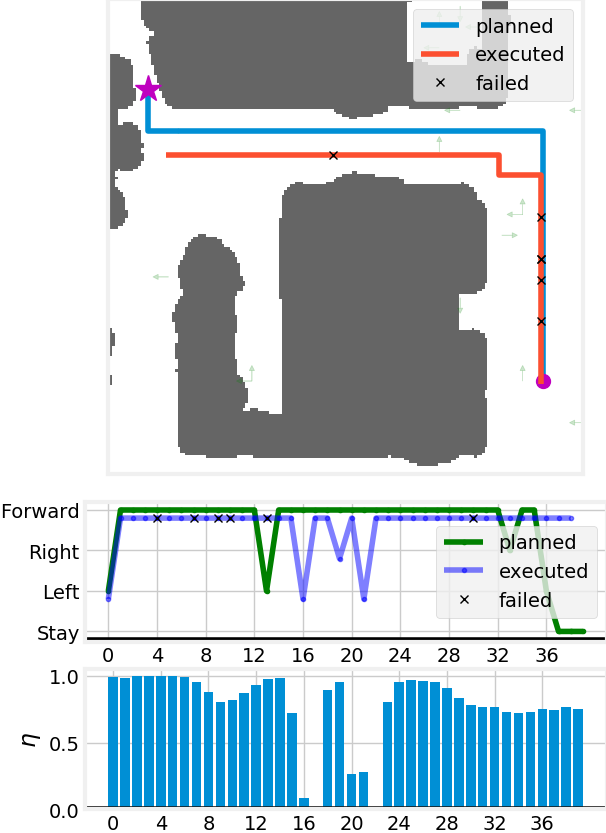} \hfill \hfill
\insertH{0.25}{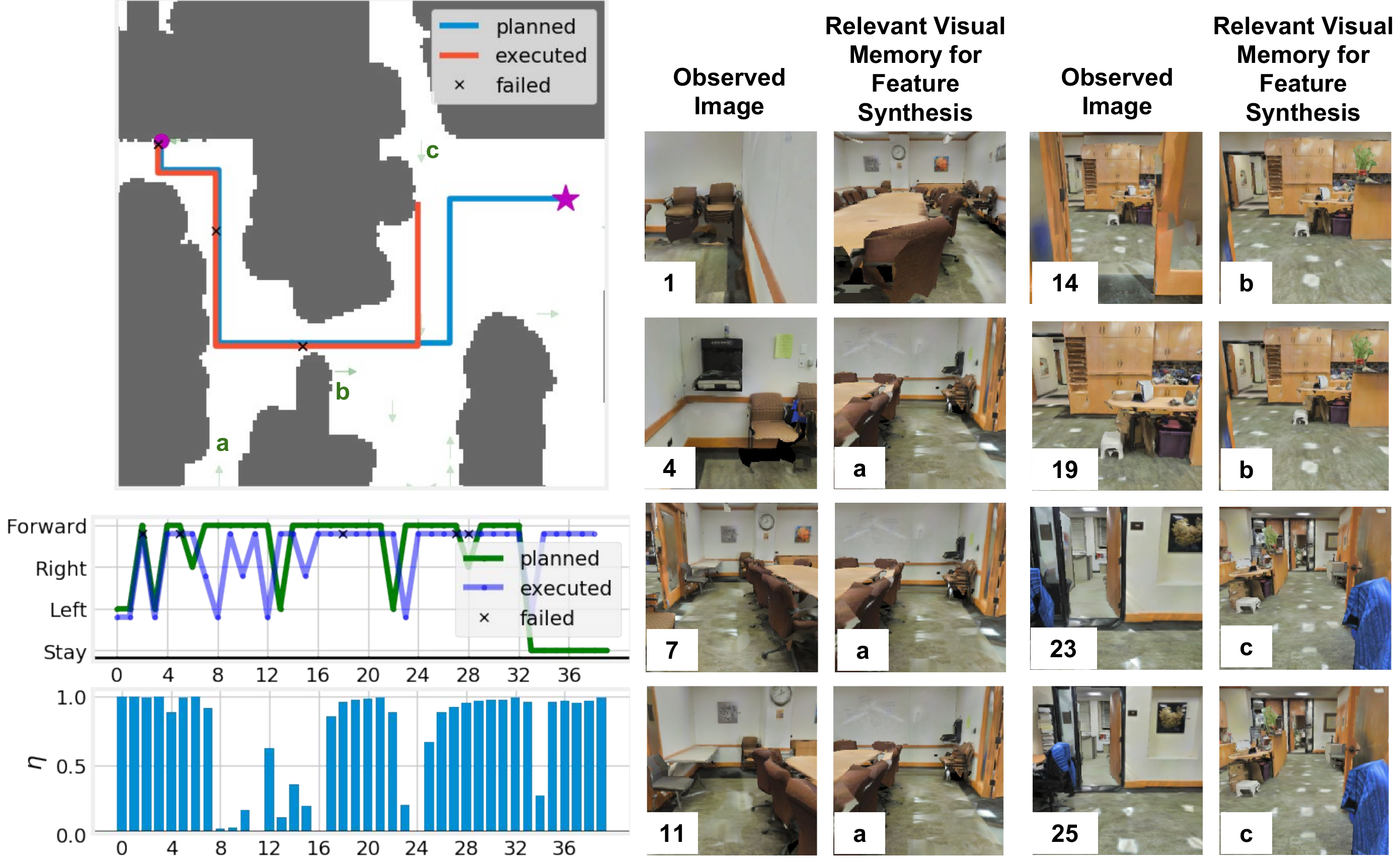} 
\caption{\textbf{Visualization of Executed Trajectories:} We show some example
executions for our method for the case where the goal is 32 to 36 time steps
away and only random images from the environment are available for reference.
We visualize the top view of the environment and display the planned and
executed trajectories. We also show the profile of actions in the optimal plan
and the executed trajectory below the map.  Failed actions due to noise in
actuation are marked in red. The bottom most plot shows the time profile for
$\eta$. The left figure shows a simple success case and shows how our closed
loop controller is able to robustly get to the goal location. The middle two
plots show some interesting examples where the policy is able to execute
intricate trajectories.  The right pane additionally shows robots instantaneous
observations at different time steps (numbered $1, 4, \ldots 25$). We also show
the most relevant reference image that was used to synthesize the visual fixes
for these steps and mark where they come from in the environment (marked with
a, b, and c). We see that the agent uses relevant images to derive the signal
for visual fixes it must use to follow the desired trajectory. Videos
visualizations are available on the project website.}
\figlabel{synth_vis_2}
\vspace{-0.1in}
\end{figure*}

\par \noindent \textit{Vanilla GRU Policy:} We also experiment with a GRU-based
policy where we replace our structured policy network with a vanilla GRU. We
extract a fixed length representation for the plan by passing it through a GRU.
This fixed length representation along with the current observation is used by
the policy GRU to output the action that the agent should take.

\paragraph{Settings.} We also study different settings, in particular, how the
visual memories ($\mathcal{I}$) were chosen. We consider two cases: (1) visual
memories are sampled from along the path that the agent needs to follow, and
(2) the more challenging case when visual memories are randomly sampled from
the environment (40 random images over an area of $8m \times 8m$ around the
trajectory). Our formulation is agnostic to the choice of these images as it
synthesizes the necessary views from whatever ones that are available.

\newcommand{\Aopen}{\phz6.2}
\newcommand{\Bopen}{\phz8.0}
\newcommand{\Copen}{26.9}
\newcommand{\Aclose}{\phz5.1}
\newcommand{\Bclose}{\phz7.0}
\newcommand{\Cclose}{35.7}
\newcommand{\AOurr}{\phz4.8}
\newcommand{\BOurr}{\phz6.0}
\newcommand{\COurr}{37.4}
\newcommand{\AOura}{\phz4.4}
\newcommand{\BOura}{\phz6.0}
\newcommand{\COura}{43.9}
\newcommand{\AOurb}{\phz3.8}
\newcommand{\BOurb}{\phz5.0}
\newcommand{\COurb}{52.7}
\newcommand{\ALSTMa}{\phz5.1}
\newcommand{\BLSTMa}{\phz6.0}
\newcommand{\CLSTMa}{36.5}
\newcommand{\ALSTMb}{\phz5.0}
\newcommand{\BLSTMb}{\phz6.0}
\newcommand{\CLSTMb}{37.9}
\newcommand{\ALSTMr}{\phz5.2}
\newcommand{\BLSTMr}{\phz7.0}
\newcommand{\CLSTMr}{35.1}
\newcommand{\ALSTMc}{\phz4.6}
\newcommand{\BLSTMc}{\phz6.0}
\newcommand{\CLSTMc}{43.2}
\newcommand{\AOurc}{\phz1.8}
\newcommand{\BOurc}{\phz3.0}
\newcommand{\COurc}{80.5}
\newcommand{\Ainit}{16.0}
\newcommand{\Binit}{17.0}
\newcommand{\Cinit}{\phz0.0}

\renewcommand{\arraystretch}{1.2} 
\renewcommand{\th}{\textsuperscript{th}} 
\setlength{\tabcolsep}{12pt}
\begin{table}
\centering
\footnotesize
\resizebox{1.0\linewidth}{!}{
\begin{tabular}{lccc}\toprule
                                            & Mean     & 75\th \%ile & Success \\
Method                                      & Distance & Distance    & \%age \\ \midrule
Initial                                     & \Ainit   & \Binit      & \Cinit \\
Open Loop Action Execution                  & \Aopen   & \Bopen      & \Copen \\
Learned Action Execution                    & \Aclose  & \Bclose     & \Cclose \\
\midrule
 GRU (5 from path)                  & \ALSTMa  & \BLSTMa     & \CLSTMa \\
 GRU (10 from path)                 & \ALSTMb  & \BLSTMb     & \CLSTMb \\
 GRU (all from path)                & \ALSTMc  & \BLSTMc     & \CLSTMc \\
 GRU (40 from env)                  & \ALSTMr  & \BLSTMr     & \CLSTMr \\
\midrule
Our (5 from path)                   & \AOura   & \BOura      & \COura \\
Our (10 from path)                  & \AOurb   & \BOurb      & \COurb \\
Our (all from path)                 & \AOurc   & \BOurc      & \COurc \\
Our (40 from env)                   & \AOurr   & \BOurr      & \COurr \\
\bottomrule 
\end{tabular}}
\caption{\textbf{Navigation Results:} We report metrics based on distance to
goal after policy execution for 20 steps. See text for details.}
\tablelabel{policy-eval}
\vspace{-0.2in}
\end{table}

\paragraph{Results.} Results are reported in \tableref{policy-eval}.  The agent
starts 16 steps away from the goal on average. Open loop execution of action
takes the agent to 6.2 steps away, succeeding about 27\% of the times, showing
that simple replay of actions fails under noisy actuation. Learning to execute
actions in context of the current observation performs much better, boosting
the success rate to 35.7\%. This shows the benefit of allowing the policy to
adjust according to the observed images.  Incorporating visual memory further
boosts performance, both when using a vanilla GRU and our proposed
architecture. When all visual memories along the path are available, our
proposed policy has a success rate of 80\% compared to 43\% for the vanilla GRU
policy. This shows the effectiveness of our proposed architecture over a
vanilla GRU. We also studied a setting where images being input to the policy
were perturbed (color changes, small affine transforms), while keeping the
images used to generate the path signatures fixed. We saw only a minor
degradation in performance (73\% vs 80\% success), showing that such learned
policies can be trained to be robust to changes in environment.

We also implemented a nearest neighbor approach that computes similarity of the
current observation to views in the path signature. This similarity is then
used to pick the optimal action. All operations are implemented in a
differentiable manner such that the feature space for nearest neighbor matching
can be learned. This nearest neighbor approach worked very well when given {\it
all} images along the path as input (success rate of 84\%), but dramatically
failed when only a subset of images were available (success rate between 5\%
and 12\%). 

\newcommand{\Aallopennonoise}{\phz3.80}
\newcommand{\Ballopennonoise}{\phz8.00}
\newcommand{\Callopennonoise}{69.53}
\newcommand{\Aallopen}{\phz7.61}
\newcommand{\Ballopen}{11.00}
\newcommand{\Callopen}{20.34}
\newcommand{\Aallclose}{\phz6.93}
\newcommand{\Ballclose}{10.00}
\newcommand{\Callclose}{26.65}
\newcommand{\AallOurr}{\phz6.86}
\newcommand{\BallOurr}{\phz9.00}
\newcommand{\CallOurr}{26.90}
\renewcommand{\Ainit}{15.99}
\renewcommand{\Binit}{17.00}
\renewcommand{\Cinit}{\phz0.00}

\renewcommand{\arraystretch}{1.2} 
\renewcommand{\th}{\textsuperscript{th}} 
\setlength{\tabcolsep}{8pt}
\begin{table}
\centering
\footnotesize
\resizebox{1.0\linewidth}{!}{
\begin{tabular}{lccc}\toprule
                                     & Mean             & 75\th \%ile      & Success \\
Method                               & Distance         & Distance         & \%age \\ \midrule
Initial                              & \Ainit           & \Binit           & \Cinit \\
Open Loop Plan Execution             & \Aallopen        & \Ballopen        & \Callopen \\
Learned Plan Execution               & \Aallclose       & \Ballclose       & \Callclose \\
Our                                  & \AallOurr        & \BallOurr        & \CallOurr \\ \midrule
Open Loop Plan Execution (No Noise)  & \Aallopennonoise & \Ballopennonoise & \Callopennonoise \\
\bottomrule
\end{tabular}}
\caption{\textbf{Full System Results:} We report performance of our full system
for navigation with noisy actuation. We report metrics based on distance to
goal after policy execution for 20 steps. See text for details.}
\tablelabel{policy-eval-full}
\vspace{-0.2in}
\end{table}

When we vary the number of visual memories given, performance gracefully
degrades. In the extreme case, when no visual memories on the planned path are
available (denoted `40 from env'), our method still outperforms the no visual
memory baseline. We also present the distribution of the final distance to goal
over different trajectories in \figref{loc_ap}~(far right) for a much harder
test case of following longer trajectories (goal between 32 and 36 time steps
away). We notice consistent improvements over open loop execution, as well as,
learned closed loop policy. \figref{synth_vis_2} shows some policy executions,
policy execution videos are available on the project website.

\noindent \textbf{Full System Evaluation.} We also evaluate the performance of
the full system in \tableref{policy-eval-full} in the most challenging setting
where the agent is given 40 reference images from the environment, and it has
to get to a specified goal location (sampled to be between 16 and 18 steps
away), under noisy actuation. We use the joint mapper and planner to compute a
plan, that is executed by the learned closed-loop controller. Planning succeeds
in 70\% cases, open loop execution of these plans succeeds in 20\% cases, and
our learned plan execution with visual memories succeeds about 27\% of the
time.

\noindent \textbf{Discussion.} This work presents some steps towards learned
mapping and landmark based representations for visual navigation. Future work
should study more intimate interaction between mapping and landmark systems
where path planning can itself be guided by distinctiveness of landmarks along
the way, study the problem in continuous action spaces and conduct experiments
on real robots.

\noindent \textbf{Acknowledgements:}
This work was supported in part by Intel/NSF VEC award IIS-1539099, and the
Google Fellowship to Saurabh Gupta.

{\small
\bibliographystyle{ieee}
\bibliography{refs-vision,refs-da,refs-robo,refs-nav}

\begin{thebibliography}{10}\itemsep=-1pt

\bibitem{matterport}
Matterport.
\newblock \url{https://matterport.com/}.

\bibitem{armeni20163d}
I.~Armeni, O.~Sener, A.~R. Zamir, H.~Jiang, I.~Brilakis, M.~Fischer, and
  S.~Savarese.
\newblock {3D} semantic parsing of large-scale indoor spaces.
\newblock In {\em CVPR}, 2016.

\bibitem{axelrod2017provably}
B.~Axelrod, L.~P. Kaelbling, and T.~Lozano-P{\'e}rez.
\newblock Provably safe robot navigation with obstacle uncertainty.
\newblock In {\em RSS}, 2017.

\bibitem{brahmbhatt2017deepnav}
S.~Brahmbhatt and J.~Hays.
\newblock Deepnav: Learning to navigate large cities.
\newblock In {\em CVPR}, 2017.

\bibitem{canny1988complexity}
J.~Canny.
\newblock {\em The complexity of robot motion planning}.
\newblock MIT press, 1988.

\bibitem{daftry2016learning}
S.~Daftry, J.~A. Bagnell, and M.~Hebert.
\newblock Learning transferable policies for monocular reactive mav control.
\newblock In {\em ISER}, 2016.

\bibitem{davison1998mobile}
A.~J. Davison and D.~W. Murray.
\newblock Mobile robot localisation using active vision.
\newblock In {\em ECCV}, 1998.

\bibitem{doucet2000rao}
A.~Doucet, N.~De~Freitas, K.~Murphy, and S.~Russell.
\newblock Rao-blackwellised particle filtering for dynamic bayesian networks.
\newblock In {\em UAI}, 2000.

\bibitem{engel2014lsd}
J.~Engel, T.~Sch{\"o}ps, and D.~Cremers.
\newblock {LSD-SLAM}: Large-scale direct monocular {SLAM}.
\newblock In {\em ECCV}, 2014.

\bibitem{slam-survey:2015}
J.~Fuentes-Pacheco, J.~Ruiz-Ascencio, and J.~M. Rend\'{o}n-Mancha.
\newblock Visual simultaneous localization and mapping: a survey.
\newblock {\em Artificial Intelligence Review}, 2015.

\bibitem{giusti2016machine}
A.~Giusti, J.~Guzzi, D.~C. Cire{\c{s}}an, F.-L. He, J.~P. Rodr{\'\i}guez,
  F.~Fontana, M.~Faessler, C.~Forster, J.~Schmidhuber, G.~Di~Caro, et~al.
\newblock A machine learning approach to visual perception of forest trails for
  mobile robots.
\newblock {\em RAL}, 2016.

\bibitem{gupta2017cognitive}
S.~Gupta, J.~Davidson, S.~Levine, R.~Sukthankar, and J.~Malik.
\newblock Cognitive mapping and planning for visual navigation.
\newblock In {\em CVPR}, 2017.

\bibitem{izadiUIST11}
S.~Izadi, D.~Kim, O.~Hilliges, D.~Molyneaux, R.~Newcombe, P.~Kohli, J.~Shotton,
  S.~Hodges, D.~Freeman, A.~Davison, and A.~Fitzgibbon.
\newblock {KinectFusion}: real-time {3D} reconstruction and interaction using a
  moving depth camera.
\newblock {\em UIST}, 2011.

\bibitem{jaderberg2015spatial}
M.~Jaderberg, K.~Simonyan, A.~Zisserman, et~al.
\newblock Spatial transformer networks.
\newblock In {\em NIPS}, 2015.

\bibitem{kahn2017self}
G.~Kahn, A.~Villaflor, B.~Ding, P.~Abbeel, and S.~Levine.
\newblock Self-supervised deep reinforcement learning with generalized
  computation graphs for robot navigation.
\newblock {\em arXiv preprint arXiv:1709.10489}, 2017.

\bibitem{kavraki1996probabilistic}
L.~E. Kavraki, P.~Svestka, J.-C. Latombe, and M.~H. Overmars.
\newblock Probabilistic roadmaps for path planning in high-dimensional
  configuration spaces.
\newblock {\em RA}, 1996.

\bibitem{khan2017memory}
A.~Khan, C.~Zhang, N.~Atanasov, K.~Karydis, V.~Kumar, and D.~D. Lee.
\newblock Memory augmented control networks.
\newblock {\em arXiv preprint arXiv:1709.05706}, 2017.

\bibitem{kingma2014adam}
D.~Kingma and J.~Ba.
\newblock Adam: A method for stochastic optimization.
\newblock {\em arXiv preprint arXiv:1412.6980}, 2014.

\bibitem{klein2007parallel}
G.~Klein and D.~Murray.
\newblock Parallel tracking and mapping for small {AR} workspaces.
\newblock In {\em ISMAR}, 2007.

\bibitem{konolige2010view}
K.~Konolige, J.~Bowman, J.~Chen, P.~Mihelich, M.~Calonder, V.~Lepetit, and
  P.~Fua.
\newblock View-based maps.
\newblock {\em IJRR}, 2010.

\bibitem{lavalle2000rapidly}
S.~M. Lavalle and J.~J. Kuffner~Jr.
\newblock Rapidly-exploring random trees: Progress and prospects.
\newblock In {\em Algorithmic and Computational Robotics: New Directions},
  2000.

\bibitem{levine2016end}
S.~Levine, C.~Finn, T.~Darrell, and P.~Abbeel.
\newblock End-to-end training of deep visuomotor policies.
\newblock {\em JMLR}, 2016.

\bibitem{mirowski2016learning}
P.~Mirowski, R.~Pascanu, F.~Viola, H.~Soyer, A.~Ballard, A.~Banino, M.~Denil,
  R.~Goroshin, L.~Sifre, K.~Kavukcuoglu, et~al.
\newblock Learning to navigate in complex environments.
\newblock In {\em ICLR}, 2017.

\bibitem{nister2004visual}
D.~Nist{\'e}r, O.~Naroditsky, and J.~Bergen.
\newblock Visual odometry.
\newblock In {\em CVPR}, 2004.

\bibitem{parisotto2017neural}
E.~Parisotto and R.~Salakhutdinov.
\newblock Neural map: Structured memory for deep reinforcement learning.
\newblock {\em arXiv preprint arXiv:1702.08360}, 2017.

\bibitem{pomerleau1989alvinn}
D.~A. Pomerleau.
\newblock Alvinn: An autonomous land vehicle in a neural network.
\newblock In {\em NIPS}, 1989.

\bibitem{ross2011reduction}
S.~Ross, G.~J. Gordon, and D.~Bagnell.
\newblock A reduction of imitation learning and structured prediction to
  no-regret online learning.
\newblock In {\em AISTATS}, 2011.

\bibitem{sadeghi2016cadrl}
F.~Sadeghi and S.~Levine.
\newblock {(CAD)$^2$RL}: Real singel-image flight without a singel real image.
\newblock In {\em RSS}, 2017.

\bibitem{schaal1997learning}
S.~Schaal.
\newblock Learning from demonstration.
\newblock In {\em NIPS}, 1997.

\bibitem{siciliano2016springer}
B.~Siciliano and O.~Khatib.
\newblock {\em Springer handbook of robotics}.
\newblock Springer, 2016.

\bibitem{tamar2016value}
A.~Tamar, S.~Levine, and P.~Abbeel.
\newblock Value iteration networks.
\newblock In {\em NIPS}, 2016.

\bibitem{thrun2005probabilistic}
S.~Thrun, W.~Burgard, and D.~Fox.
\newblock {\em Probabilistic robotics}.
\newblock MIT press, 2005.

\bibitem{tolman1948cognitive}
E.~C. Tolman.
\newblock Cognitive maps in rats and men.
\newblock {\em Psychological review}, 1948.

\bibitem{wiener2009taxonomy}
J.~M. Wiener, S.~J. B{\"u}chner, and C.~H{\"o}lscher.
\newblock Taxonomy of human wayfinding tasks: A knowledge-based approach.
\newblock {\em Spatial Cognition \& Computation}, 9(2):152--165, 2009.

\bibitem{zhang2017neural}
J.~Zhang, L.~Tai, J.~Boedecker, W.~Burgard, and M.~Liu.
\newblock Neural slam.
\newblock {\em arXiv preprint arXiv:1706.09520}, 2017.

\bibitem{zhou2017unsupervised}
T.~Zhou, M.~Brown, N.~Snavely, and D.~G. Lowe.
\newblock Unsupervised learning of depth and ego-motion from video.
\newblock In {\em CVPR}, 2017.

\bibitem{zhu2016target}
Y.~Zhu, R.~Mottaghi, E.~Kolve, J.~J. Lim, A.~Gupta, L.~Fei-Fei, and A.~Farhadi.
\newblock Target-driven visual navigation in indoor scenes using deep
  reinforcement learning.
\newblock In {\em ICRA}, 2017.

\end{thebibliography}
}
\end{document}